\pdfoutput=1

\documentclass[11pt]{article}

\usepackage[]{acl}

\usepackage{times}
\usepackage{latexsym}

\usepackage[T1]{fontenc}

\usepackage[utf8]{inputenc}

\usepackage{tcolorbox}
\usepackage{colortbl}
\usepackage{tabularx}
\usepackage{tabu}
\usepackage{booktabs}
\usepackage{multirow}

\usepackage{graphicx} 
\usepackage{float}
\usepackage{subfigure} 
\usepackage{amssymb}

\usepackage{hyperref}
\usepackage{tablefootnote}
\usepackage{xspace}

\usepackage{float}
\usepackage{amsmath}
\usepackage{bm}
\usepackage{algorithmicx}
\usepackage{algorithm}
\usepackage{algpseudocode}

\usepackage{arydshln}

\usepackage{amssymb}
\usepackage{pifont}
\newcommand{\cmark}{\ding{51}}%
\newcommand{\xmark}{\ding{55}}%
\usepackage{enumitem}

\usepackage{microtype}

\newcommand{\ours}{\textsc{SciVer}\xspace}
\newcommand{\eg}{\hbox{\emph{e.g.,}}\xspace}
\newcommand{\nexample}{3,000\xspace}
\newcommand{\npaper}{1,113\xspace}
\newcommand{\nmodel}{21\xspace}
\newcommand{\ie}{\hbox{\emph{i.e.,}}\xspace}
\newcommand{\entail}{\emph{``entailed''}\xspace}
\newcommand{\refute}{\emph{``refuted''}\xspace}

\newcommand{\huggingface}{\raisebox{-1.5pt}{\includegraphics[height=1.05em]{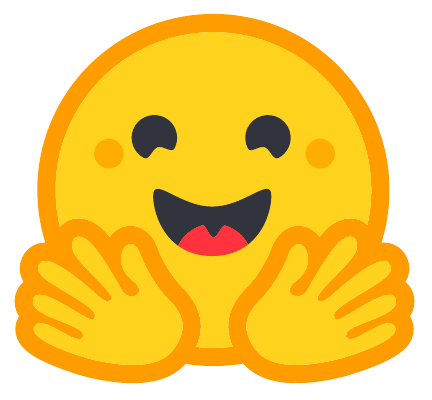}}\xspace}
\newcommand{\github}{\raisebox{-1.5pt}{\includegraphics[height=1.05em]{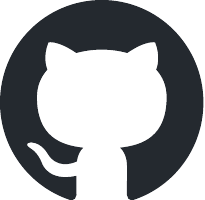}}\xspace}

\title{\ours: Evaluating Foundation Models for Multimodal \\Scientific Claim Verification}

\author{
Chengye Wang \quad
Yifei Shen \quad
Zexi Kuang \quad
Arman Cohan \quad
Yilun Zhao \vspace{10pt} \\
Yale NLP Lab
}

\begin{document}
\maketitle
\begin{abstract}
We introduce \ours, the first benchmark specifically designed to evaluate the ability of foundation models to verify claims within a multimodal scientific context.
\ours consists of 3,000 expert-annotated examples over 1,113 scientific papers, covering four subsets, each representing a common reasoning type in multimodal scientific claim verification. 
To enable fine-grained evaluation, each example includes expert-annotated supporting evidence.
We assess the performance of \nmodel state-of-the-art multimodal foundation models, including o4-mini, Gemini-2.5-Flash, Llama-3.2-Vision, and Qwen2.5-VL. 
Our experiment reveals a substantial performance gap between these models and human experts on \ours.
Through an in-depth analysis of retrieval-augmented generation (RAG), and human-conducted error evaluations, we identify critical limitations in current open-source models, offering key insights to advance models' comprehension and reasoning in multimodal scientific literature tasks.

\begin{small}
\begin{center}
\begin{tabular}{ll}
\huggingface~~\textbf{Data} &  \href{https://huggingface.co/datasets/chengyewang/SciVer} {\path{chengyewang/SciVer}}\\
\github~~\textbf{Code} &\href{https://github.com/QDRhhhh/SciVer}{\path{QDRhhhh/SciVer}}\\
\end{tabular}
\end{center} 
\end{small}
\end{abstract}

\section{Introduction}
\begin{figure*}[!t]
\centering
\includegraphics[width=1\linewidth]{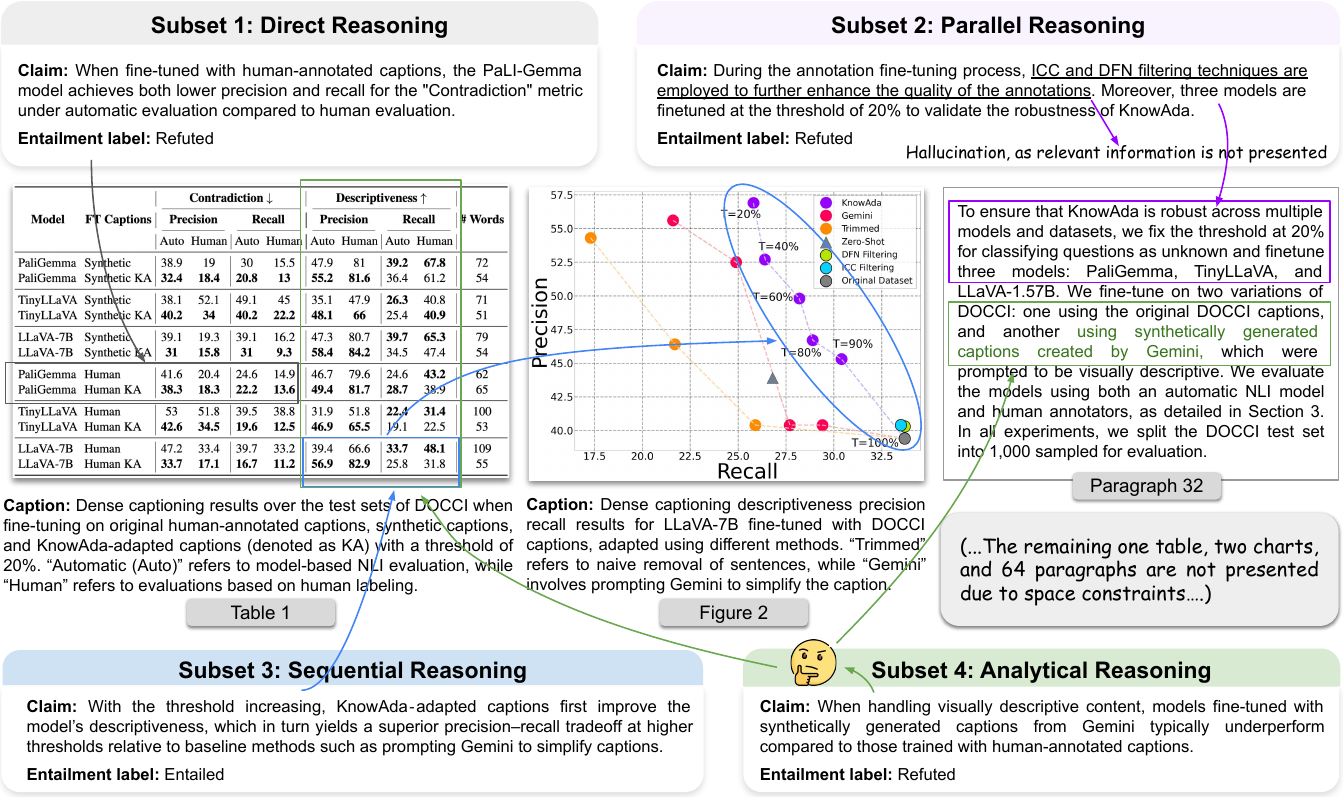}
\caption{
An illustration of the four subsets in the \ours benchmark. Our benchmark is designed to evaluate document-grounded scientific claim verification in a multimodal setting. To effectively perform this task, models must go through the full context of a scientific paper—including text, charts, and tables—to locate the appropriate supporting evidence before verifying a claim. The complete data examples are provided in Appendix~\ref{app:err}.
}
\label{fig:main-example}
\end{figure*}
Scientific claim verification has become increasingly vital as the research community grapples with an ever-expanding body of scientific literature across diverse domains~\cite{dasigi-etal-2021-dataset, wadden-etal-2022-scifact, qasa, asai2024openscholar}. The accuracy of claim verification in a scientific paper is not merely a matter of cross-checking numerical consistency or validating conclusions—it necessitates a holistic understanding of the paper’s context (\eg textual content, charts, and tables).

Despite the significance of multimodal reasoning, existing benchmarks in scientific claim verification have often treated these components in isolation. Predominantly, prior works have focused either on textual content alone~\cite{wadden-etal-2022-scifact} or on verifying claims based on a single table~\cite{lu-etal-2023-scitab}. While previous multimodal question-answering (QA) benchmarks in scientific literature comprehension incorporate scientific charts, they still remain limited to QA tasks over a single chart~\cite{mmsci, charxiv, li-etal-2024-multimodal-arxiv}, failing to capture the broader multimodal context of scientific literature.
Consequently, the lack of a comprehensive multimodal benchmark restricts the systematic evaluation of foundation models' ability to reason across the diverse and interconnected modalities in scientific literature.

In this work, we introduce \textbf{\ours}, 
a comprehensive and high-quality benchmark for evaluating multimodal \textbf{\textsc{Sci}}entific claim \textbf{\textsc{Ver}}ification.
\ours consists of \nexample expert-annotated examples over \npaper scientific papers spanning diverse domains within computer science. 
To ensure our benchmark reflects real-world scenarios in scientific literature comprehension, we design four fine-grained tasks (as illustrated in \autoref{fig:main-example}): \emph{direct reasoning}, \emph{parallel reasoning}, \emph{sequential reasoning}, and \emph{analytical reasoning}. 
Each task targets a common reasoning type in multimodal scientific claim verification.
Moreover, each example includes expert-annotated supporting evidence, facilitating fine-grained performance evaluation.

We conduct an extensive evaluation on \ours, covering \nmodel frontier open-source and proprietary multimodal foundation models. 
Our experimental results reveal that while state-of-the-art models achieve human-comparable performance on simpler tasks (\eg direct reasoning), 
they continue to struggle with more complex challenges. For instance, GPT-4.1, achieves an accuracy of 70.8\% on analytical reasoning, falling significantly short of human expert performance (\ie 90.0\%). 
This demonstrates the challenging nature of \ours.
Furthermore, our analysis of retrieval-augmented generation (RAG) and human-conducted error analyses provide insights for future advancement.

We summarize our contributions as follows:
\begin{itemize} [leftmargin=*]
\itemsep0em 
\item We introduce a new claim verification benchmark to challenge foundation models across diverse reasoning scenarios in multimodal scientific literature comprehension. Each example undergoes expert annotation and strict quality control to ensure benchmark reliability and high standards.
\item We conduct an extensive evaluation that encompasses \nmodel open-source and proprietary foundation models, comprehensively assessing their capabilities and limitations in our task.
\item We provide an in-depth analysis of Chain-of-Thought reasoning, RAG settings, and model reasoning errors, offering valuable insights for future advancements and targeted improvements.

\end{itemize}
\begin{table*}[!t]
\centering
\renewcommand{\arraystretch}{1.05}
\resizebox{\textwidth}{!}{%
\addtolength{\tabcolsep}{-0.35em}
\begin{tabular}{lllcc}
\toprule
\multirow{2}{*}{\textbf{Dataset}} 
& \multirow{2}{*}{\textbf{Input Context}} 
& \multirow{2}{*}{\textbf{Data Construction}} 
&  \multirow{2}{*}{\textbf{\begin{tabular}[c]{@{}c@{}}\# Task /\\Subsets\end{tabular}}}
& \multirow{2}{*}{\textbf{\begin{tabular}[c]{@{}c@{}}Rationale\\Annotation?\end{tabular}}} \\ 
&\\
\midrule
\multicolumn{5}{c}{\textbf{\textit{Scientific Literature Comprehension}}} \\
\textsc{Qasper}~\cite{dasigi-etal-2021-dataset} & Single NLP paper & Expert annotation & 4 & Evidence \\
QASA~\cite{qasa} & Single AI/ML paper (text-only) & Expert annotation & 3 & Evidence \\
MMSci~\cite{mmsci} & Multiple figures or charts from STEM papers & GPT-4o generation & 2 & \xmark \\
ArXivQA~\cite{li-etal-2024-multimodal-arxiv} & Single chart from arXiv papers & GPT-4V generation& -- & \xmark \\
CharXiv~\cite{charxiv} & Single Chart from arXiv papers & Expert annotation & 2 & \xmark \\
\textsc{SciFact}~\cite{wadden-etal-2020-fact} & Multiple STEM paper abstracts & Expert annotation & -- & --\\

\noalign{\vskip 0.5ex}\hdashline\noalign{\vskip 0.5ex}
\multicolumn{5}{c}{\textbf{\textit{Claim Verification over Multimodal Context}}} \\
\textsc{InfoTabS}~\cite{gupta-etal-2020-infotabs} & Single wikipedia table & Crowdsourcing & -- & \xmark \\
\textsc{TabFact}~\cite{Chen2020TabFact} & Single wikipedia table & Crowdsourcing & 2 & \xmark \\
ChartCheck~\cite{akhtar-etal-2024-chartcheck} & Single wikipedia chart & Crowdsourcing & 2 & Rationale \\
\textsc{SciTAB}~\cite{lu-etal-2023-scitab} & Single scientific table from NLP\&ML paper & Expert+InstructGPT & -- & \xmark \\
\midrule
\textbf{\ours} (ours) & Multiple tables, charts, paragraphs from CS papers & Expert annotation & 4 & Evidence\\
\bottomrule
\end{tabular}
}
\caption{
Comparison of \ours with existing claim verification and scientific literature comprehension benchmarks.
}
\label{tab:dataset_comparison}
\end{table*}
\section{Related Work}
\subsection{Claim Verification}
Claim verification is a well-established research area that can be categorized into two main settings. The first is the open-domain setting, where an external retriever is used to fetch relevant information from a large corpus to verify claims~\cite{vlachos-riedel-2014-fact, thorne-etal-2018-fever, aly-etal-2021-fact, wadden-etal-2022-scifact, rangapur2024finfactbenchmarkdatasetmultimodal}. 
The second is context-grounded claim verification, where claims are verified based solely on given context, without relying on external retrieval~\cite{Chen2020TabFact, kamoi-etal-2023-wice, lu-etal-2023-scitab, glockner-etal-2024-ambifc, zhao-etal-2024-findver}.
This work focuses on the latter setting, as it removes variability introduced by retriever performance and enables a more controlled evaluation of foundation models' ability to verify claims within multimodal scientific context. 
As shown in \autoref{tab:dataset_comparison}, existing multimodal claim verification benchmarks primarily use either a single table~\cite{Chen2020TabFact, gupta-etal-2020-infotabs, lu-etal-2023-scitab} or single chart~\cite{akhtar-etal-2024-chartcheck} as input context. 
In real-world scenarios, however, verifying claims in scientific literature requires reasoning across multiple modalities, including textual descriptions, tables, and figures. 

\subsection{Scientific Literature Comprehension}
With the rapid expansion of research publications, evaluating and applying foundation models for scientific literature comprehension has become increasingly important~\cite{asai2024openscholar, skarlinski2024languageagentsachievesuperhuman, li-etal-2024-m3sciqa}. 
Existing benchmarks primarily focus on QA tasks, assessing models on their ability to extract or infer information from scientific papers~\cite{dasigi-etal-2021-dataset, qasa}. 
While recent efforts have extended QA tasks to incorporate tabular and visual information~\cite{li-etal-2024-multimodal-arxiv, charxiv, mmsci}, they remain constrained by their single-modality focus, neglecting the rich multimodal context inherent in scientific papers.
Claim verification, on the other hand, demands a more comprehensive understanding of scientific literature, as claims are often supported by a combination of textual descriptions, tables, and charts. 
Additionally, each example in \ours includes detailed supporting evidence, facilitating fine-grained evaluation.
\section{\ours Benchmark}\label{sec:data}
\begin{figure*}[!t]
 \centering
\includegraphics[width=\textwidth]{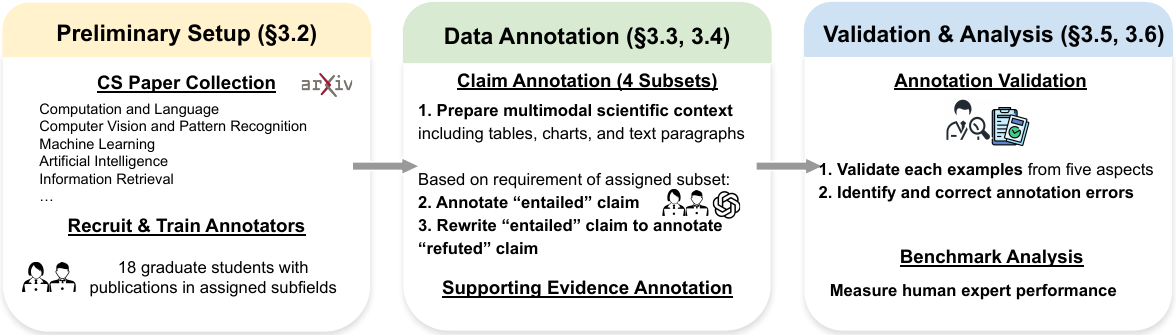}
 \caption{An overview of the \ours benchmark construction pipeline.}
 \label{fig:data_construction_pipeline}
\end{figure*}

\ours is a comprehensive evaluation framework designed to assess the ability of foundation models to verify scientific claims within a multimodal context. \autoref{fig:data_construction_pipeline} provides an overview of the \ours construction pipeline. 
In the following subsections, we detail the benchmark design, data construction process, and quality validation methodology.

\subsection{Benchmark Design}
We first present the task formulation and the four specialized subsets of our dataset that we designed to evaluate different aspects of model performance.

\paragraph{Task Formulation.}
We formally define the task of \ours within the context of a foundation model $FM$ as follows:
Given a scientific claim $c$ and multimodal contexts $\{P, I, T\}$ collected from a scientific paper—where $P$ denotes textual paragraphs, $I$ denotes multiple charts, and $T$ denotes multiple tables—the model is is tasked with determining the entailment label $\ell \in \mathcal{L} = \{\text{\entail},\text{\refute}\}$:

\begin{equation}
\label{eq:labeling}
    \ell = \arg\max_{\ell \in \mathcal{L}} P_{~\mathbf{FM}}(\ell~|~c, P, I, T)
\end{equation}

\noindent It challenges foundation models to perform complex reasoning by integrating and interpreting textual, tabular, and visual data to verify scientific claims. 
Since scientific tables often have intricate structures that are difficult to represent in textual format, we follow recent work in multimodal table understanding~\cite{zheng-etal-2024-multimodal, deng-etal-2024-tables} by using table screenshots as inputs.

\paragraph{Subset Design.}
\ours includes the following four distinct subsets, each designed to evaluate a specific reasoning type commonly required for scientific claim verification over multimodal context:

\noindent{(1) \emph{Direct Reasoning}},
which evaluates models' ability to extract and interpret a single piece of information to verify a scientific claim. 

\noindent{(2) \emph{Parallel Reasoning}}, which evaluates models' ability to simultaneously process and integrate information from multiple distinct sources. 

\noindent{(3) \emph{Sequential Reasoning}}, which evaluates models' ability to perform step-by-step inference chains across different modalities. Models are required to establish logical connections between multiple pieces of evidence, where each step's conclusion becomes a premise for subsequent reasoning steps. 

\noindent{(4) \emph{Analytical Reasoning}}, which evaluates models' ability to verify claims that require both sophisticated domain knowledge and complex reasoning beyond direct data extraction. Models must not only interpret the provided data but also apply relevant scientific principles and methodological understanding to arrive at valid conclusions.

\noindent Appendix~\ref{app:err} presents detailed examples of each subset. These subsets enable fine-grained evaluation across different reasoning paradigms commonly encountered in scientific literature comprehension.

\subsection{Preliminary Setup}
We next discuss the preliminary setup for data construction, including the process of scientific paper collection and expert annotator recruitment.
\paragraph{Expert Annotator Recruitment and Training.}
Existing claim verification datasets primarily rely on crowdsourced data curation (as shown in \autoref{tab:dataset_comparison}). 
However, our preliminary study suggests that crowd-sourced annotators often lack the necessary domain expertise for our task. To mitigate this, we recruit 18 CS graduate students with relevant subject-specific knowledge, requiring each to have at least two peer-reviewed publications in their assigned subfields. 
Detailed annotator biographies are provided in \autoref{app:annotator} in Appendix.
To further enhance annotation quality and consistency, all selected experts undergo a mandatory two-hour individual training session with one of the authors, ensuring that they are familiar with the annotation guidelines and protocol.

\paragraph{Scientific Paper Collection.}
\ours focuses on arXiv papers published between September 1, 2024, and November 15, 2024, covering eight key areas of \emph{computer science}. 
To ensure high-quality content, we prioritize papers that include comments indicating acceptance by a peer-reviewed venue. 
For each paper, we extract its multimodal context—including textual content, tables, and charts—from the HTML versions available on the arXiv platform. 
We filter out papers that contain fewer than two tables or two charts.

\subsection{Claim Annotation}\label{sec:claim}
Given a paper relevant to their research field, the annotators follow these steps for claim annotation:
\paragraph{Multimodal Scientific Context Preparation.}
Scientific papers are often lengthy, exceeding the maximum context length of certain foundation models. Including the full text may overwhelm these models and hinder their ability to integrate information effectively across modalities. 
To address this, annotators refine the paper context by removing textual sections that are not essential to understanding the core research problem, such as related work, acknowledgments, references, and appendix sections.

\paragraph{Entailed Claim Annotation.}
To reduce bias stemming from the positioning of evidence, the annotation interface randomly selects three charts or tables from the curated context, along with their surrounding textual paragraphs. 
Annotators are then tasked with writing an entailed claim that aligns with the pre-given reasoning types (\ie subset). They are required to ensure that verifying the claim requires referencing at least one of the three sampled multimodal elements. Subsequently, annotators identify all relevant supporting evidence, which is later reviewed by a second annotator.

\paragraph{Refuted Claim Annotation.}
Following established practices in the field~\cite{wadden-etal-2022-scifact, Chen2020TabFact, lu-etal-2023-scitab}, and given the difficulty of directly obtaining \refute claims, we instead generate them by perturbing original \entail claims through a semi-automated annotation process.
Specifically,
to curate \refute claims, annotators modify the initially annotated \entail claim by introducing factual errors that contradict the supporting evidence.

\subsection{Supporting Evidence Annotation}\label{sec:rationale}
After completing the claim annotation, a second annotator, who is also an expert in the relevant research field, is tasked with annotating the supporting evidence. 
The annotators are required to carefully review the claim and identifying all relevant paragraphs, tables, and charts that serve as supporting evidence. 
To ensure consistency and accuracy, we compare the supporting evidence and entailment label annotated in this step with those from the initial claim annotation. If discrepancies arise between the two annotations, a third expert annotator is introduced to adjudicate the differences. 
Our process achieves an \emph{inter-annotator agreement} of 94.0\% for entailment label annotation, demonstrating strong reliability in our annotation.









\begin{table}[!t]
\footnotesize
\centering
\renewcommand{\arraystretch}{1.05}
\begin{tabular}{lrr}
\toprule
\textbf{Property} (avg.) & \textbf{Val} & \textbf{Test} \\
\midrule
\multicolumn{3}{c}{\textbf{\textit{Multimodal Scientific Context}}} \\
\noalign{\vskip 0.5ex}

\# Words in text paragraphs & 583.6 & 567.4 \\
\# Tables & 0.55 & 0.54\\
\quad Table caption length & 14.2 & 13.7 \\
\# Charts & 0.94 & 0.95\\
\quad Chart caption length & 39.2 & 40.2 \\
\midrule

\multicolumn{3}{c}{\textbf{\textit{Claim Verification}}} \\
\noalign{\vskip 0.5ex}
Claim length & 30.5 &  33.9 \\
\quad \# Entailed & 505 & 995\\
\quad \# Refuted & 495 & 1,005 \\
\noalign{\vskip 0.5ex}\hdashline\noalign{\vskip 0.5ex}
Supporting Evidence & 2.63 & 2.62\\

\midrule
Scientific papers & 327 & 786 \\
Total examples & 1,000 & 2,000\\
\bottomrule
\end{tabular}

\caption{Data statistics of \ours.}
\label{tab:data-stat}
\end{table}
\subsection{Data Validation}
Each annotated example undergoes a comprehensive validation process conducted by a different expert annotator within the same research field. The validation focuses on the following five aspects:
(1) The claim must be grammatically correct, well-structured, and free of spelling or typographical errors.
(2) The claim must align with the annotation requirements of its corresponding subset and should not be verifiable using textual context alone.
(3) The claim must be meaningfully situated within the paper context and hold practical significance for scientific literature comprehension.
(4) The annotated supporting evidence must be directly relevant to the claim and comprehensive enough to support claim verification without requiring additional, unannotated context.

If an example fails to meet any of these criteria, validators are responsible for making necessary revisions. In practice, 232 initially annotated examples required revisions before being finalized. 

\subsection{Data Statistics and Analysis} 
\autoref{tab:data-stat} presents the data statistics of \ours. 
It is randomly divided into the validation and test sets. 
The validation set contains 1,000 examples and is intended for model development and validation. 
The test set comprises the remaining 2,000 examples and is designed for standard evaluation.

To approximate \textbf{human-expert-level performance} on \ours, we randomly sampled 10 claims from each subset, totaling 40 claims. Two expert annotators independently evaluated these claims, providing the natural language explanation and final entailment label for each claim. They achieve an average accuracy of 93.8\% (\autoref{tab:main_results}).

\begin{figure}[!t]
\begin{tcolorbox}[colback=black!3!white, colframe=black!70!white, title=Adopted Chain-of-Thought Prompt, fontupper=\footnotesize, fonttitle=\footnotesize]

\textcolor{blue}{\{Paper Context (textual paragraphs, tables, charts)\}} \\

You are given a multimodal scientific context that includes textual paragraphs, tables, and charts. Your task is to determine whether the given claim is Entailed or Refuted. Be skeptical and cautious: if there is any inconsistency, missing evidence, or ambiguity, consider the claim incorrect.\\

Claim to verify:\\
\textcolor{blue}{\{Claim\}}
\newline\newline
Start by explaining your reasoning process clearly, focusing on identifying potential contradictions, lack of support, or misleading interpretations. Think step by step before answering.

\end{tcolorbox}
\caption{The Chain-of-Thought prompt used.
}\label{fig:cot-prompt}
\end{figure}
\begin{table*}[!t]
\centering
\renewcommand\arraystretch{1.1} 
\footnotesize
\begin{tabular}{ll*{4}{>{\centering\arraybackslash}p{1.1cm}}*{2}{>{\centering\arraybackslash}p{1.1cm}}}
\toprule[1pt]
 & \multirow{3}{*}{\textbf{Release}} &\multicolumn{4}{c}{\textbf{Test Set}} 
& \multirow{2}{*}{\textbf{\begin{tabular}[c]{@{}c@{}}Avg. \\  Validation\end{tabular}} } & \multirow{2}{*}{\textbf{\textbf{\begin{tabular}[c]{@{}c@{}}Avg. \\  Test\end{tabular}} }} \\
\cline{3-6}\noalign{\vskip 1ex}
 & & \textbf{Direct} & \textbf{Parallel} & \textbf{Sequential} & \textbf{Analytical} \\
\midrule[0.7pt]
\multicolumn{8}{c}{\textbf{\textit{Baseline Settings}}} \\
\noalign{\vskip 1ex}
Human Expert & & 100.0 & 95.0 & 90.0 & 90.0 & -- & 93.8\\
Random Guess & & 50.0 & 50.0 & 50.0 & 50.0 & 50.0 & 50.0\\

\midrule
\multicolumn{8}{c}{\textbf{\textit{Proprietary Models}}} \\
\noalign{\vskip 1ex}
o4-mini & 2025-04 & \cellcolor{red!35}{85.0} & \cellcolor{red!35}{80.6} & \cellcolor{red!35}{77.6} & 67.6 & \cellcolor{red!35}{79.6} & \cellcolor{red!35}{77.7} \\
Gemini-2.5-Flash & 2025-05 & \cellcolor{red!20}{79.8} & \cellcolor{red!20}{76.0} & \cellcolor{red!5}{73.2} & 71.4 & \cellcolor{red!20}{76.0} & \cellcolor{red!20}{75.1} \\
GPT-4o & 2024-11 & 77.0 & 71.2 & \cellcolor{red!20}{73.6} & \cellcolor{red!20}{73.8} & 72.3 & \cellcolor{red!5}{73.9} \\
Gemini-2.0-Flash & 2025-02 & \cellcolor{red!5}{78.0} & 72.2 & 69.4 & \cellcolor{red!5}{73.4} & 73.0 & 73.3 \\
GPT-4.1 & 2025-04 & 77.6 & \cellcolor{red!5}{73.2} & 71.2 & 70.8 & \cellcolor{red!5}{74.3} & 73.2 \\
GPT-4o-mini & 2024-07 & 71.4 & 67.6 & 61.4 & 62.0 & 63.8 & 65.6 \\

\midrule
\multicolumn{8}{c}{\textbf{\textit{Open-source Models}}} \\
\noalign{\vskip 1ex}
Mistral-Small-3.1-24B & 2025-03 & 74.8 & 66.0 & 68.6 & \cellcolor{red!35}{75.6} & 73.6 & 71.3 \\
Qwen2.5-VL-72B & 2025-01 & 70.8 & 69.2 & 68.2 & 69.2 & 70.2 & 69.4 \\
InternVL3-38B & 2025-04 & 65.8 & 64.6 & 65.2 & 70.4 & 70.6 & 66.5 \\
Qwen2-VL-72B & 2024-11 & 70.4 & 61.0 & 63.0 & 67.2 & 65.9 & 65.4 \\
InternVL2.5-38B & 2024-11 & 65.0 & 55.8 & 62.4 & 66.8 & 63.8 & 62.5 \\
Pixtral-12b & 2024-09 & 60.8 & 54.6 & 63.4 & 65.2 & 61.1 & 61.0 \\
InternVL3-8B & 2025-04 & 64.2 & 54.6 & 56.0 & 63.0 & 58.8 & 59.5 \\
Qwen2.5-VL-7B & 2025-01 & 55.8 & 57.4 & 57.0 & 60.2 & 53.5 & 57.6 \\
InternVL2.5-8B & 2024-11 & 53.8 & 56.4 & 53.2 & 58.2 & 55.5 & 55.4 \\
InternVL2-8B & 2024-06 & 54.0 & 52.6 & 50.2 & 54.6 & 52.9 & 52.9 \\
Qwen2-VL-7B & 2024-11 & 52.6 & 54.0 & 52.0 & 52.0 & 52.8 & 52.7 \\
Llama-3.2-11B-Vision & 2024-09 & 53.6 & 50.6 & 51.8 & 53.2 & 48.9 & 52.3 \\
Phi-4-Multimodal & 2025-03 & 50.8 & 50.8 & 51.2 & 51.0 & 52.1 & 51.0 \\
LLaVA-OneVision & 2024-09 & 49.8 & 48.2 & 49.6 & 53.6 & 51.0 & 50.3 \\
Phi-3.5-Vision & 2024-08 & 46.0 & 52.0 & 48.0 & 49.2 & 51.5 & 48.8 \\

\bottomrule[1pt]
\end{tabular}

\caption{
Model accuracy on \ours validation and test sets with CoT prompts, ranked by test set performance.
}
\label{tab:main_results}
\end{table*}

\section{Experiment}
This section first outlines the experiment setup, and then discusses our experiment results and analysis.
\subsection{Experiment Setup}
We use accuracy as the primary metric to evaluate model performance on \ours. 
Following recent benchmark studies~\cite{yue2024mmmuprorobustmultidisciplinemultimodal, yue2025mmmupro}, we adopt rule-based methods to derive the final entailment label from the model response, which is then compared to the ground-truth label.

We evaluate a broad range of frontier foundation models that support \emph{multiple images} and text as input. 
Specifically, we evaluate \textbf{11 series of open-source models}, including 
InternVL-2, 2.5, and 3~\cite{chen2023internvl, chen2024far,chen2024expanding}, 
Qwen2-VL and Qwen2.5-VL~\cite{Qwen-VL, Qwen2VL}, 
Pixtral~\cite{agrawal2024pixtral12b},
Mistral-Small-3.1~\cite{mistral2025small31},
LLaVA-OneVision~\cite{li2024llavaonevisioneasyvisualtask},
Llama-3.2-Vision~\cite{grattafiori2024llama3herdmodels},
Phi-3.5-Vision and Phi-4-Multimodal ~\cite{abdin2024phi3,microsoft2025phi4}.
We also evaluate \textbf{five series of proprietary models}, including 
OpenAI o4-mini~\cite{openai2025o4mini}, GPT-4o and GPT-4.1 ~\cite{openai2024gpt4o,openai2024gpt4-1}, 
Gemini-2.0 and Gemini-2.5~\cite{geminiteam2024gemini, geminiteam2025gemini}.
Appendix \ref{app:model_info} details the parameter settings and configurations of the evaluated models.
For open-source models, we utilize the vLLM pipeline~\cite{kwon2023efficient} for model inference; while for proprietary models, we use their official API service.

We evaluate the models with the \textbf{Chain-of-Thought} prompt, which is presented in \autoref{fig:cot-prompt}.

\begin{figure}[!t]
 \centering
\includegraphics[width=\linewidth]{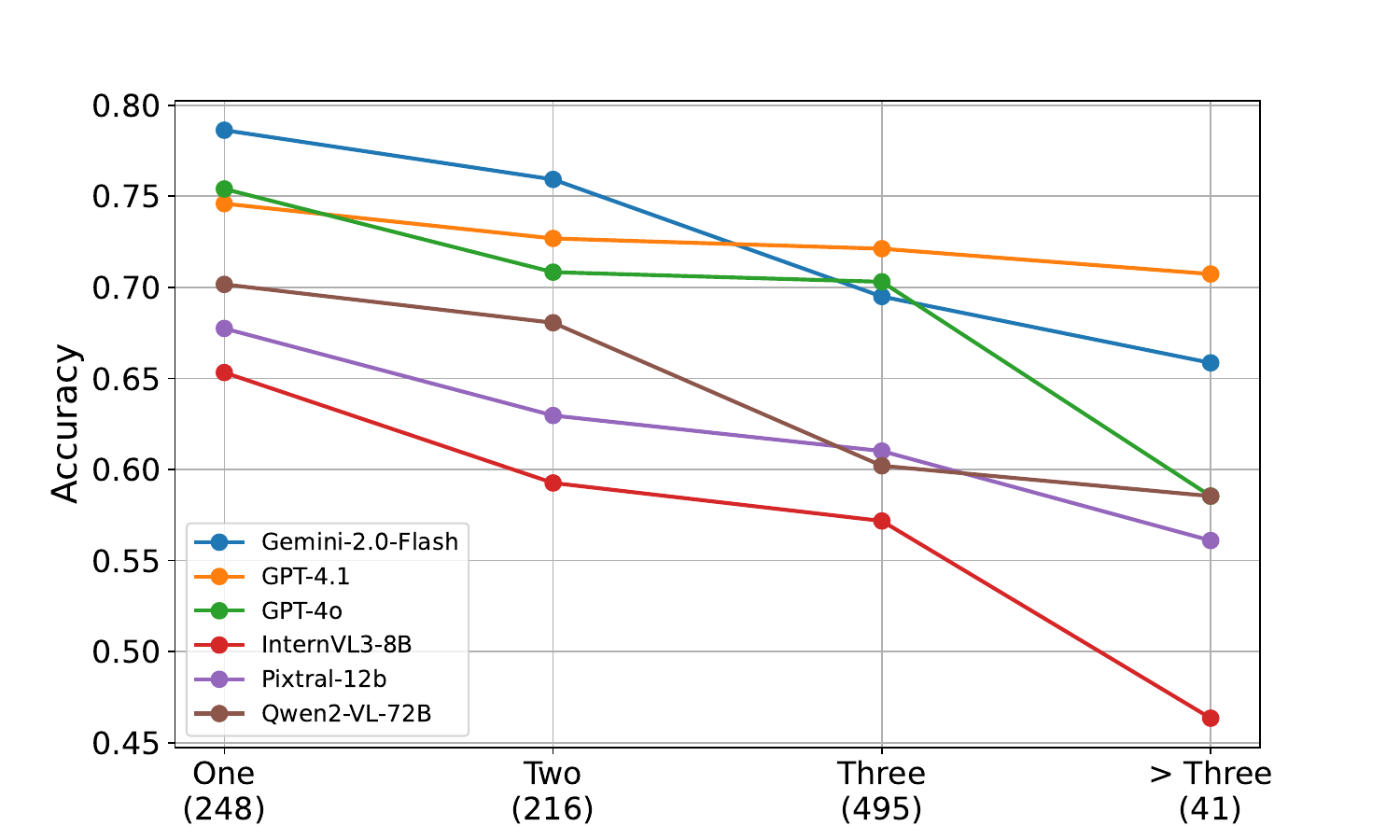}
 \caption{Comparison of model performance on the validation set, with claims requiring varying amounts of annotated supporting evidence. Each piece of evidence is defined as a single table, chart, or paragraph (\S\ref{sec:rationale}).
 }
 \label{fig:line_chart}
\end{figure}
\begin{figure*}[!t]
\centering
\includegraphics[width=1\linewidth]{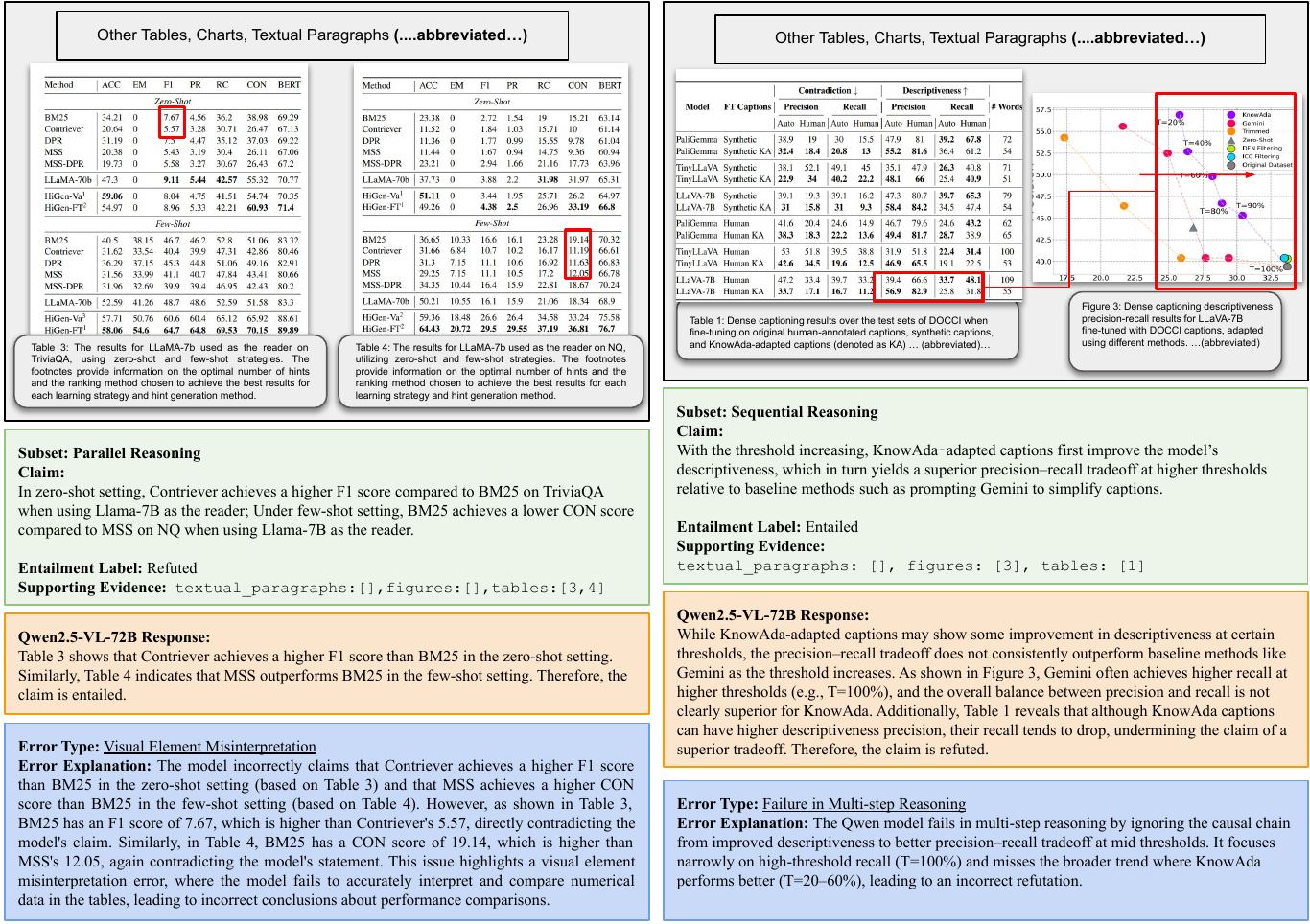}
\caption{
Illustration of two error types: Visual Element Misinterpretation (left) and Failure in Multi-step Reasoning (right). Additional error examples are provided in Appendix~\ref{app:err}.
}
\label{fig:error_analysis}
\end{figure*}
\subsection{Main Findings}
\autoref{tab:main_results} presents the evaluated models’ performance. Our main findings are as follows:
\paragraph{\ours presents substantial challenges for current models.} 
While the recently released reasoning models, o4-mini and Gemini-2.5-Flash, demonstrate leading performance, other models fall short of human expert capabilities.
For instance, GPT-4.1 achieves 73.2\% accuracy with CoT prompting, considerably lower than the 93.8\% accuracy achieved by human experts.
This performance gap highlights \ours’s crucial role in advancing and assessing the capabilities of models in multimodal scientific literature comprehension.

\paragraph{Performance of open-sourced models.}
Open-source models continue to lag behind their proprietary counterparts. However, models such as Mistral-Small-3.1, Qwen2.5-VL, and InternVL3 have achieved competitive performance, narrowing the gap with top proprietary models. These advancements highlight the rapid progress in open-source development. In the following subsections, we provide a detailed analysis of open-source models and offer insights for future improvements.

\paragraph{Model performance declines with increasing evidence requirements.}
To provide a fine-grained analysis of model performance on multi-hop reasoning in \ours, we compare frontier models on the validation set across claims that require different numbers of annotated supporting evidence.
As shown in \autoref{fig:line_chart}, model performance consistently declines as the number of ground-truth evidence pieces increases. This trend suggests that current models struggle with multi-hop reasoning and with synthesizing information across multiple multimodal contexts.

\subsection{Error Analysis and Case Study}
To better understand the limitations of open-source models, we perform a detailed error analysis on Qwen2.5-VL-72B. We randomly select 25 instances from each of the four subsets for evaluation.
Through a detailed inspection of model response, we identify five common error types:

\begin{itemize} [leftmargin=*]
\itemsep0em 
\item \textbf{Failure to Retrieve Relevant Information} (32\%), where models fail to retrieve and consider all the key evidence from the provided multimodal context, leading to incomplete reasoning or incorrectly classify verifiable claims as lacking enough information.
\item \textbf{Visual element misinterpretation} (21\%), where models misinterpret charts or tables.
\item \textbf{Failure in multi-step reasoning} (17\%), where models struggle to connect intermediate reasoning steps over extracted information, leading to incorrect entailment predictions.
\item \textbf{Heavy reliance on text modality} (12\%), where models focus primarily on textual input, failing to properly integrate crucial information from tables and charts.
\item \textbf{Domain-specific misconceptions} (10\%), where models misapply domain terminology or rely on irrelevant memorized knowledge when verifying the given claims.
\item \textbf{Other observed errors} include incorrect numerical computations and instances where models refuse to generate a response. 
\end{itemize}

For each error type, we provide illustrative examples and corresponding error analyses in \autoref{fig:error_analysis} and Appendix~\ref{app:err}.

\subsection{Retrieval-Augmented Generation Analysis}\label{sec:rag}
The preceding error analysis highlights that the failure to retrieve relevant information is a primary error type. This finding motivates us to explore how RAG settings can be leveraged to improve model performance on \ours. 

\begin{figure}[!t]
\begin{tcolorbox}[colback=black!3!white, colframe=black!70!white, title=Prompt for Evidence Filtering, fontupper=\footnotesize, fonttitle=\footnotesize]

\textcolor{blue}{\{Single Multimodal Element\}} \\

Analyze the given context and determine whether it contains relevant information to verify the following claim: \textcolor{blue}{\{Claim\}}
\newline
Respond with either ``yes'' if the context contains the necessary information to verify the claim, or ``no'' if it does not.

\end{tcolorbox}
\caption{
The prompt for evidence filtering in \S\ref{sec:rag}.
}\label{fig:evidence}
\end{figure}
\paragraph{Experiment Setup.} 
Implementing RAG for scientific multimodal data presents challenges, as 
existing open-source retrieval models do not natively support scientific tables and charts. To overcome this limitation, we construct the textual representations for tables and charts as the \emph{concatenation of their original captions and GPT-4o-generated descriptions}. 
Each representation is indexed as separate evidence alongside the textual paragraphs extracted from the paper.
We evaluate three widely used retrieval systems, \ie BM25, Contriever~\cite{izacard2021contriever}, and OpenAI’s text-embedding-3-large, to retrieve the top-$5$ most relevant evidence for the given claim. 
The retrieved evidence is then fed into the model in its original form.
Additionally, we assess an alternative setting (\ie Evidence Filtering) where the model first determines, one by one, whether each piece of evidence is relevant to the claim (prompt shown in \autoref{fig:evidence}), and then incorporates all confirmed relevant evidence into the final input.

\begin{table}[!t]
\centering
\small
\addtolength{\tabcolsep}{-0.32em}
\begin{tabular}{lccc}
\toprule
\textbf{Setting} & \textbf{Recall@5} & \textbf{4o-mini} & \textbf{Qwen2.5-VL} \\
\midrule
Original & -- & 63.8 & 70.2 \\
\midrule
with RAG \\
\quad Contriever & 70.7 & 64.7 \textsuperscript{↑0.9} & 71.8 \textsuperscript{↑1.6} \\
\quad BM25 & 74.3 & 65.4 \textsuperscript{↑1.6} & 72.2 \textsuperscript{↑2.0} \\
\quad OAI Embedding & 81.0 & 67.0 \textsuperscript{↑3.2} & 72.9 \textsuperscript{↑2.7} \\
\quad Oracle & -- & 73.3 \textsuperscript{↑9.5} & 75.3 \textsuperscript{↑5.1} \\
\noalign{\vskip 0.5ex}\hdashline\noalign{\vskip 0.5ex}
LLM Evidence Filter & -- & 67.5 \textsuperscript{↑3.7} & 74.4 \textsuperscript{↑4.2} \\
\bottomrule
\end{tabular}
\caption{Performance comparison of GPT-4o-mini and Qwen2.5-VL-72B under different RAG settings.}
\label{tab:rag}
\end{table}
\paragraph{Findings.}
We evaluate the GPT-4o-mini and Qwen2.5-VL-72B models on the validation set. As shown in \autoref{tab:rag}, enhancements in information retrieval quality generally lead to improved entailment classification performance on \ours. Among the three retrievers tested, the OpenAI embedding model achieves the highest retrieval accuracy, which correlates with the most substantial gains in downstream LLM performance (\ie 70.2\% $\rightarrow$ 75.3\% for Qwen2.5-VL-72B). Additionally, applying an LLM-based evidence filter further boosts overall system performance.

\section{Conclusion}
This work introduces \ours, a comprehensive benchmark for evaluating multimodal scientific claim verification. By providing a diverse set of fine-grained, expert-curated examples and a reliable automated evaluation system, \ours advances the development of foundation models capable of accurately and robustly interpreting real-world scientific texts, tables, and figures.
Our experimental results expose significant performance gaps between state-of-the-art foundation models and human experts, revealing key challenges such as reasoning limitations across textual, tabular, and visual data, as well as difficulties in retrieving and integrating relevant multimodal evidence. 

\section*{Acknowledgement}
We are grateful to Google TRC program for providing computing resources and Together AI for granting LLM API credits.
\section*{Limitations}
While \ours presents a significant advancement in multimodal scientific claim verification, there are several limitations that we acknowledge, which also point to promising directions for future research.
First, \ours is primarily constructed from computer science papers sourced from arXiv, focusing on verifying claims within this discipline. While this allows us to control for domain expertise in our annotation process and ensures high-quality claim verification, it may limit the generalizability of \ours to other fields.
Second, \ours primarily focuses on claim verification over textual paragraphs, tables, and charts, as these are the most common multimodal elements in scientific literature. However, some domains rely heavily on other modalities such as equations, figures, or experimental images, which \ours does not explicitly consider in its current version.
Third, \ours relies on expert annotations with domain expertise, ensuring high-quality labels and reasoning rationales. However, this approach is labor-intensive and may not scale easily to larger datasets.

\bibliography{anthology,custom, lmm}

\appendix

\onecolumn
\section{\ours Benchmark Construction}



\begin{table*}[h]
\centering
\small
\renewcommand{\arraystretch}{1.05}
\begin{tabular}{lllcc}
\toprule
\textbf{ID} & \textbf{Biography} & \textbf{Assigned Subjects} & \textbf{\# Relevant Publications} & \textbf{Author?}\\
\midrule
1 & 2nd year PhD & Computer Vision and Pattern Recognition & 1-5 & \xmark \\
2 & Final year PhD & Computer Vision and Pattern Recognition & 5-10 & \xmark \\
3 & Postdoc & Computer Vision and Pattern Recognition & >10 & \xmark \\
4 & -- & Computation and Language & >10 & \cmark \\
5 & -- & Computation and Language & 1-5 & \cmark \\
6 & -- & Computation and Language & 1-5 & \cmark \\
7 & 3rd year PhD & Robotics & 5-10 & \xmark \\
8 & Postdoc & Robotics & >10 & \xmark \\
9 & Final year PhD & Software Engineering & 5-10 & \xmark \\
10 & Postdoc & Software Engineering & >10 & \xmark \\
11 & 2nd year PhD & Machine Learning & 1-5 & \xmark \\
12 & 4th year PhD & Machine Learning & 5-10 & \xmark \\
13 & 3rd year PhD & Artificial Intelligence & 5-10 & \xmark \\
14 & Postdoc & Artificial Intelligence & >10 & \xmark \\
15 & Master Student & Information Retrieval & 1-5 & \xmark \\
16 & 3rd year PhD & Information Retrieval & 5-10 & \xmark \\
17 & Final year PhD & Cryptography & 5-10 & \xmark \\
18 & Postdoc & Cryptography & >10 & \xmark \\
\bottomrule
\end{tabular}
\caption{
Biographies of 18 expert annotators involved in \ours construction (Author biographies are hidden to protect identity confidentiality.
}
\label{app:annotator}
\end{table*}


\clearpage
\section{Configurations of Evaluated Models}\label{app:model_info}

\begin{table*}[h]
\centering
\footnotesize
\resizebox{\textwidth}{!}{%
\begin{tabular}{lllll}
\toprule
\textbf{Organization} & \textbf{Model} & \textbf{Release} & \textbf{Version} & \textbf{\# Inference Pipeline} \\
\midrule
\multicolumn{5}{c}{\emph{\textbf{Proprietary Models}}} \\
\midrule
\multirow{3}{*}{OpenAI} & o4-mini$^*$ & 2025-04 & \texttt{o4-mini-2025-04-16} & \multirow{4}{*}{API} \\
& GPT-4.1 &  2025-04  &  \texttt{gpt-4.1-2025-04-14} & \\
& GPT-4o &  2024-08  &  \texttt{gpt-4o-2024-08-06} & \\
& GPT-4o-mini & 2024-07 & \texttt{gpt-4o-mini-2024-07-18} & \\
\noalign{\vskip 0.5ex}\hdashline\noalign{\vskip 0.5ex}
\multirow{2}{*}{Google} & Gemini-2.5-Flash & 2025-05 & \texttt{gemini-2.5-flash-preview-05-20} & \multirow{2}{*}{API} \\
& Gemini 2.0 Flash & 2024-12 & \texttt{gemini-2.0-flash-exp} & \\
\midrule
\multicolumn{5}{c}{\emph{\textbf{Open-source Multimodal Foundation Models}}} \\
\midrule
\multirow{4}{*}{Alibaba} & Qwen2.5-VL-72B & 2025-01 & \texttt{Qwen2.5-VL-72B-Instruct} & \multirow{4}{*}{vLLM} \\
& Qwen2-VL-72B & 2024-09 & \texttt{Qwen2-VL-72B-Instruct} & \\
& Qwen2.5-VL-7B & 2025-01 & \texttt{Qwen2.5-VL-7B-Instruct} & \\
& Qwen2-VL-7B & 2024-08 & \texttt{Qwen2-VL-7B-Instruct} & \\
\noalign{\vskip 0.5ex}\hdashline\noalign{\vskip 0.5ex}
\multirow{2}{*}{Mistral AI} & Mistral-Small-3.1 & 2025-03 & \texttt{Mistral-Small-3.1-24B} & \multirow{2}{*}{vLLM} \\
& Pixtral-12B & 2024-09 & \texttt{Pixtral-12B-2409}&\\
\noalign{\vskip 0.5ex}\hdashline\noalign{\vskip 0.5ex}
\multirow{5}{*}{Shanghai AI Lab} & InternVL3-38B & 2025-04 & \texttt{InternVL-3-38B} & \multirow{5}{*}{vLLM} \\
& InternVL3-8B & 2025-04 & \texttt{InternVL3-8B} & \\
& InternVL2.5-38B & 2024-11 & \texttt{InternVL2.5-38B} & \\
& InternVL2.5-8B & 2024-11 & \texttt{InternVL2.5-8B} & \\
& InternVL2-8B & 2024-06 & \texttt{InternVL2-8B} & \\
\noalign{\vskip 0.5ex}\hdashline\noalign{\vskip 0.5ex}
\multirow{1}{*}{Meta} & Llama-3.2-11B-Vision & 2024-09 & \texttt{Llama-3.2-11B-Vision-Instruct} & \multirow{1}{*}{vLLM} \\

\noalign{\vskip 0.5ex}\hdashline\noalign{\vskip 0.5ex}
\multirow{2}{*}{Microsoft} & Phi-3.5-Vision & 2024-07 & \texttt{Phi-3.5-Vision-Instruct} & \multirow{2}{*}{vLLM}\\
&Phi-4-Multimodal&2025-03&\texttt{Phi-4-Multimodal}&\\
\noalign{\vskip 0.5ex}\hdashline\noalign{\vskip 0.5ex}
\multirow{1}{*}{Llava Hugging Face} & LLaVA-OneVision-7B & 2024-09 & \texttt{llava-onevision-qwen2-7b-ov-chat-hf} & vLLM \\

\bottomrule
\end{tabular}
}
\caption{Details of the multimodal foundation models evaluated in our study. Models are organized by organization and aligned with performance data from the main text.}
\label{tab:model_configuration}
\end{table*}

\clearpage
\section{Error Analysis}\label{app:err}



\subsection{Failure to Retrieve Relevant Information}
\begin{figure}[H]
 \centering
\includegraphics[width=\textwidth]{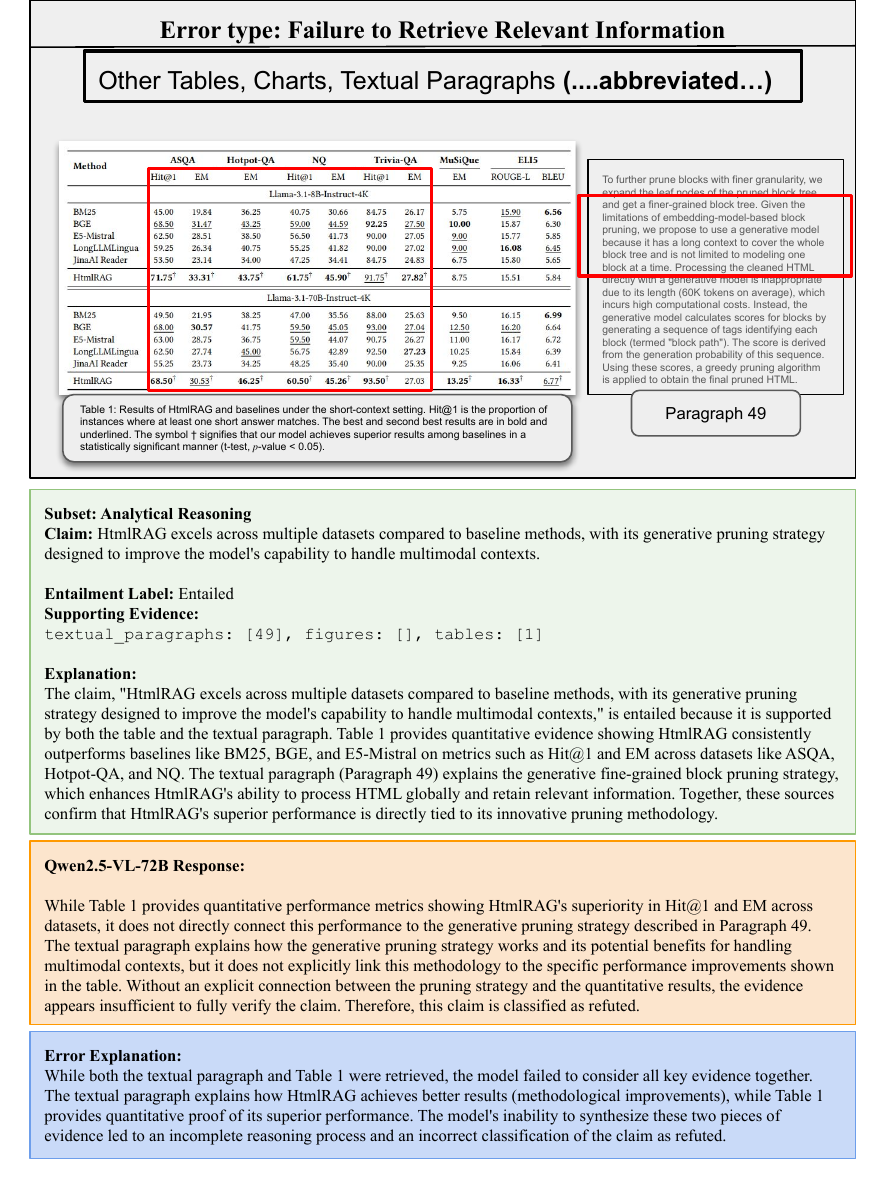}
 \caption{
Illustration of \emph{Failure to Retrieve Relevant Information} with the example from the \emph{Analytical Reasoning} subset.
 }
\end{figure}


\clearpage
\subsection{Visual element misinterpretation}
\begin{figure}[H]
 \centering
\includegraphics[width=\textwidth]{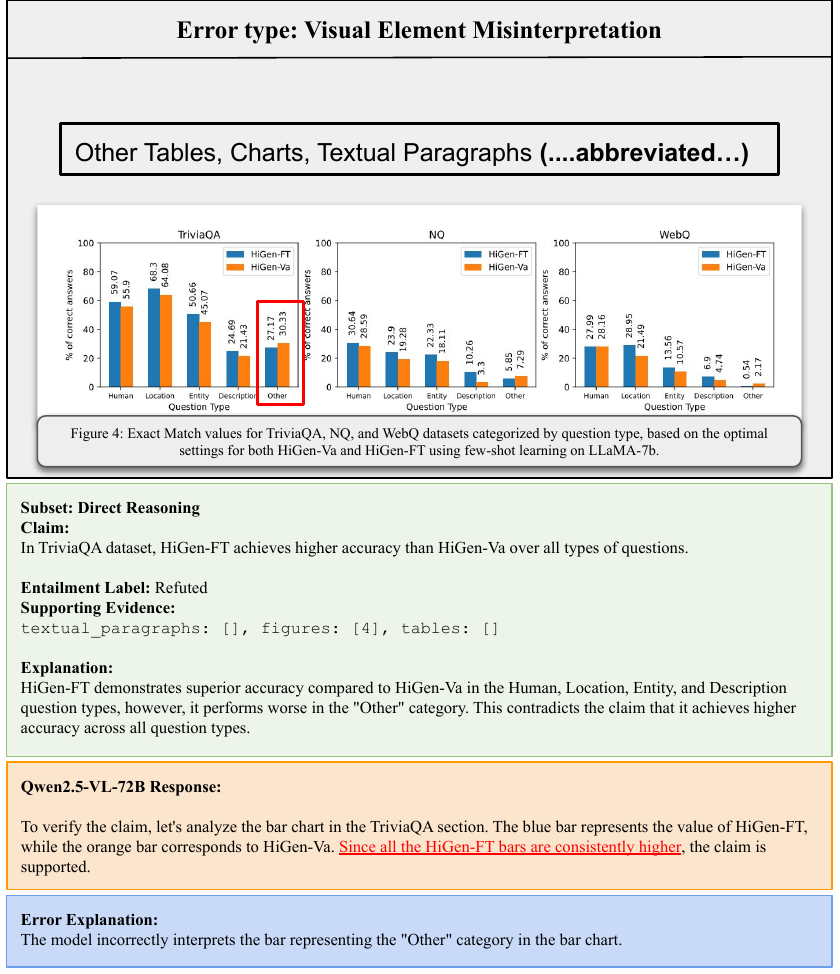}
 \caption{
Illustration of \emph{Visual element misinterpretation} with the example from the \emph{Direct Reasoning} subset.
 }
\end{figure}

\clearpage
\subsection{Heavy Reliance on Text Modality}
\begin{figure}[H]
 \centering
\includegraphics[width=\textwidth]{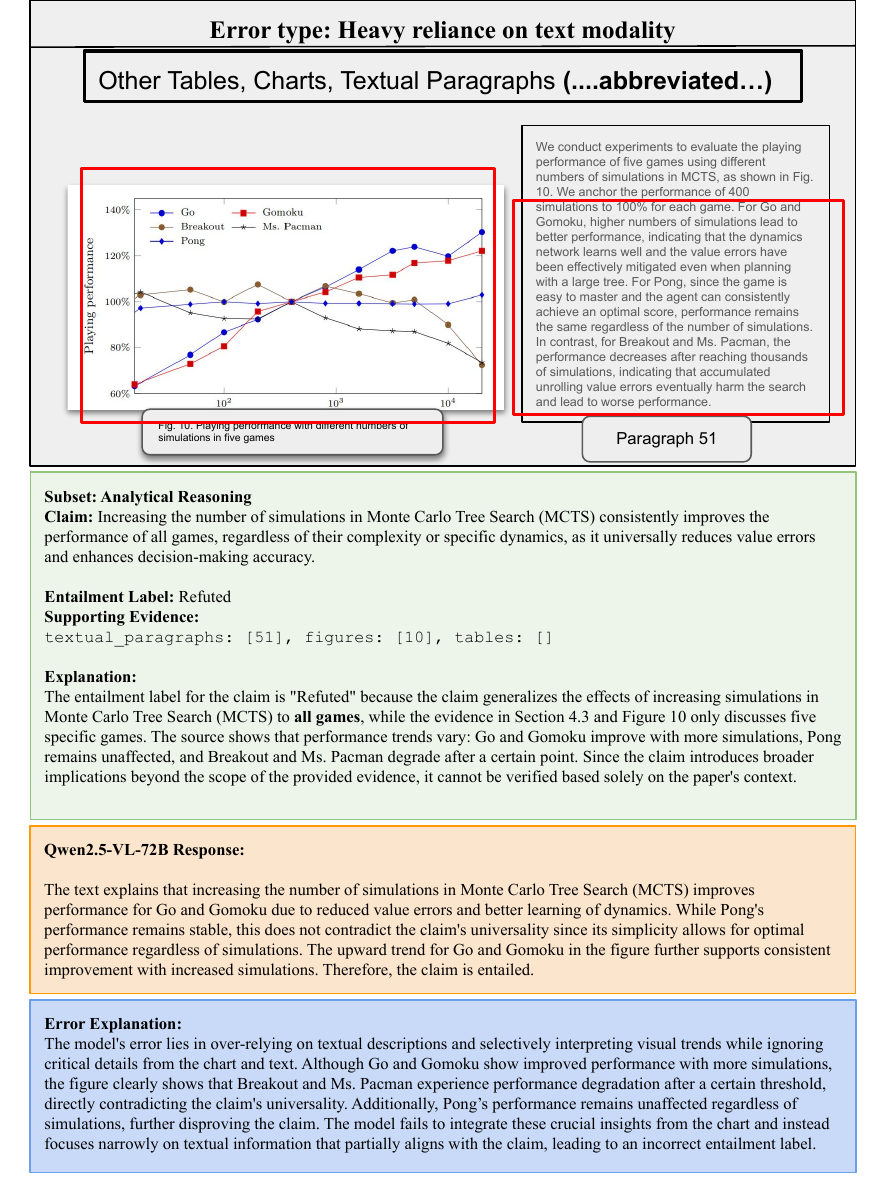}
 \caption{
Illustration of \emph{Heavy Reliance on Text Modality} with the example from the \emph{Analytical Reasoning} subset.
 }
\end{figure}

\clearpage
\subsection{Domain-Specific Misconceptions}
\begin{figure}[H]
 \centering
\includegraphics[width=\textwidth]{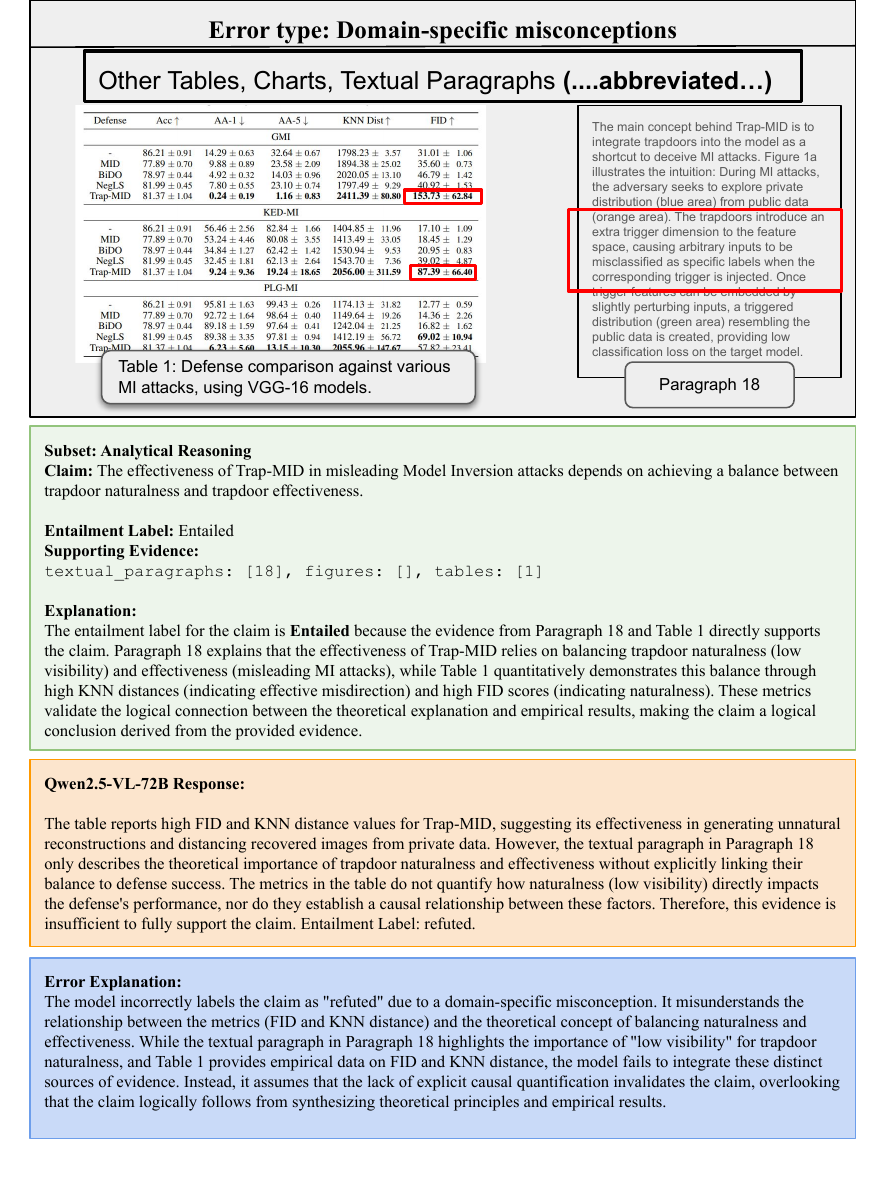}
 \caption{
Illustration of \emph{Domain-Specific Misconceptions} with the example from the \emph{Analytical Reasoning} subset.
 }
\end{figure}

\clearpage
\subsection{Other Observation Error}
\begin{figure}[H]
 \centering
\includegraphics[width=\textwidth]{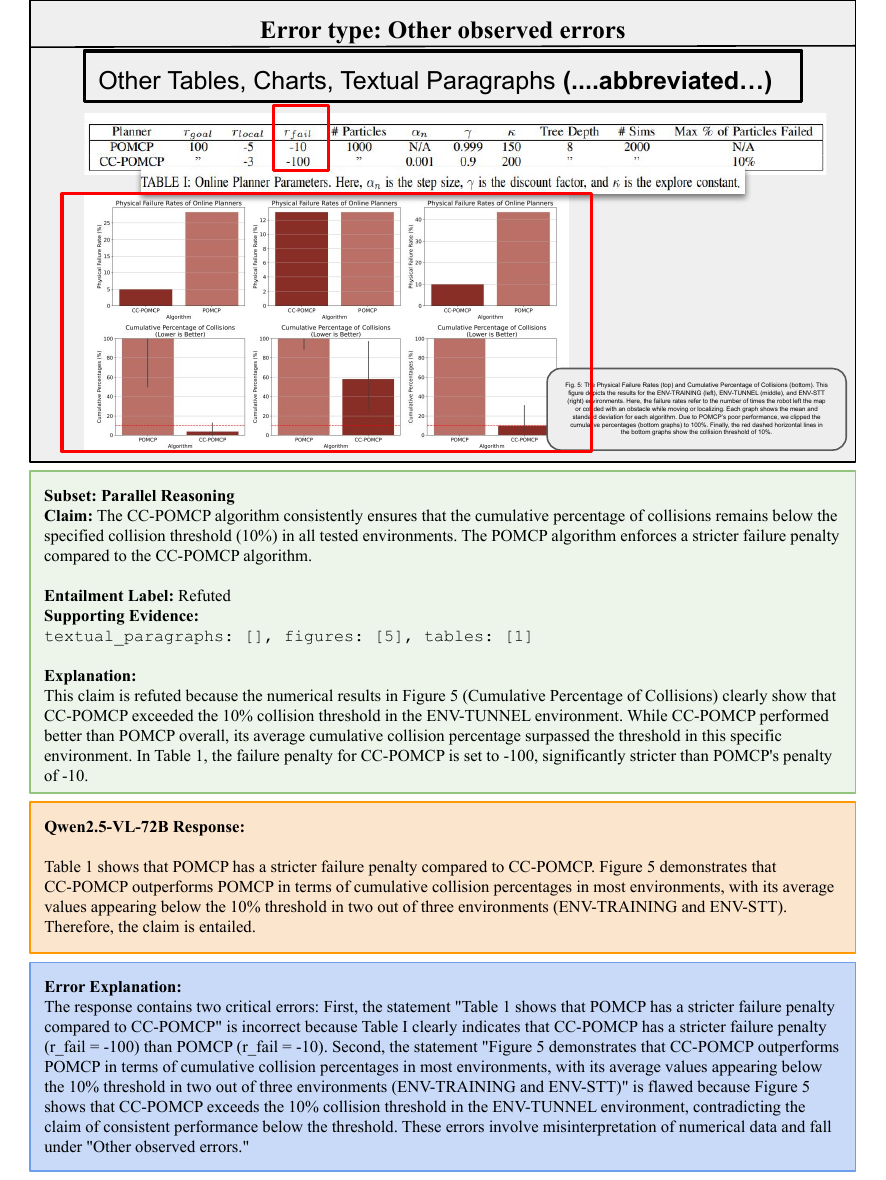}
 \caption{
Illustration of \emph{Other Observation Error} with the example from the \emph{Parallel Reasoning} subset.
 }
\end{figure}

\end{document}